\documentclass{ieeeaccess}
\usepackage{cite}
\usepackage{amsmath,amssymb,amsfonts}
\usepackage{algorithmic}
\usepackage{graphicx}
\usepackage{textcomp}
\usepackage{subfigure,cases}
\usepackage{setspace}

\usepackage{multirow}
\usepackage{caption}
\usepackage{color}
\usepackage{colortbl}
\usepackage{diagbox}

\def\BibTeX{{\rm B\kern-.05em{\sc i\kern-.025em b}\kern-.08em
    T\kern-.1667em\lower.7ex\hbox{E}\kern-.125emX}}
\begin{document}
\history{Date of publication xxxx 00, 0000, date of current version xxxx 00, 0000.}
\doi{10.1109/ACCESS.2017.DOI}

\title{Early Action Prediction with Generative Adversarial Networks}
\author{\uppercase{Dong Wang, \uppercase{Yuan Yuan}, \IEEEmembership{Senior Member, IEEE},
and Qi Wang}, \IEEEmembership{Senior Member, IEEE}}
\address{School of Computer Science and Center for OPTical IMagery Analysis and Learning (OPTIMAL),
Northwestern Polytechnical University, Xi'an 710072, China. (e-mail: nwpuwangdong@gmail.com, y.yuan1.ieee@gmail.com, crabwq@gmail.com).}

\tfootnote{
This work was supported by the National Key R\&D Program of China under Grant 2017YFB1002202, National Natural Science Foundation of China under Grant U1864204 and 61773316, State Key Program of National Natural Science Foundation of China under Grant 61632018£¬Natural Science Foundation of Shaanxi Province under Grant 2018KJXX-024, and Project of Special Zone for National Defense Science and Technology Innovation.
}

\markboth
{Author \headeretal: Preparation of Papers for IEEE TRANSACTIONS and JOURNALS}
{Author \headeretal: Preparation of Papers for IEEE TRANSACTIONS and JOURNALS}

\corresp{Corresponding author: Yuan Yuan (e-mail: y.yuan1.ieee@gmail.com).}

\begin{abstract}
Action Prediction is aimed to determine what action is occurring in a video as early as possible, which is crucial to many online applications, such as predicting a traffic accident before it happens and detecting malicious actions in the monitoring system. In this work, we address this problem by developing an end-to-end architecture that improves the discriminability of features of partially observed videos by assimilating them to features from complete videos. For this purpose, the generative adversarial network is introduced for tackling action prediction problem, which improves the recognition accuracy of partially observed videos though narrowing the feature difference of partially observed videos from complete ones. Specifically, its generator comprises of two networks: a CNN for feature extraction and an LSTM for estimating residual error between features of the partially observed videos and complete ones, and then the features from CNN adds the residual error from LSTM, which is regarded as the enhanced feature to fool a competing discriminator. Meanwhile, the generator is trained with an additional perceptual objective, which forces the enhanced features of partially observed videos are discriminative enough for action prediction. Extensive experimental results on UCF101, BIT and UT-Interaction datasets demonstrate that our approach outperforms the state-of-the-art methods, especially for videos that less than 50\% portion of frames is observed.
\end{abstract}


\titlepgskip=-15pt

\maketitle

\section{Introduction}
Humans have an enormous capacity for predicting what actions are about to happen in the near feature, which are a critical ingredient for enabling us to interact with other people effectively and avoid some dangers timely, such as cooperation between basketball players and braking before a rear-end collision. The ability to predict future actions is also a paramount technique for many components in autonomous vehicle, video surveillance and health care. For example, the intelligent driver system should predict a traffic accident before it happens, and the intelligent video surveillance system ought to alert as early as possible when malicious actions occurring. Therefore, some efforts \cite{ryoo2011human,cao2013recognize,kong2014discriminative,lan2014hierarchical} have been devoted to addressing action prediction problem, whose goal is to predict action label before the action execution ends.

Compared to the well-studied action recognition, which is to infer the action label after the entire action execution has been observed, action prediction is more challenging because the beginning of action observations from different action categories are usually ambiguous. For example, running and triple jump are similar at the beginning stage, and action prediction intends to distinguish them once the jump motion occurs. Therefore, the knowledge of difference between running and triple jump is exceedingly important, which implies that a jump motion will occur after running in triple jump action. To obtain this kind of knowledge, the videos with full action execution are necessary to learn how action appearance evolves in the temporal domain, which is the key to action prediction as it helps us to predict next movements in video concerning current observation.

\begin{figure}
  \centering
  \includegraphics[width=.45\textwidth]{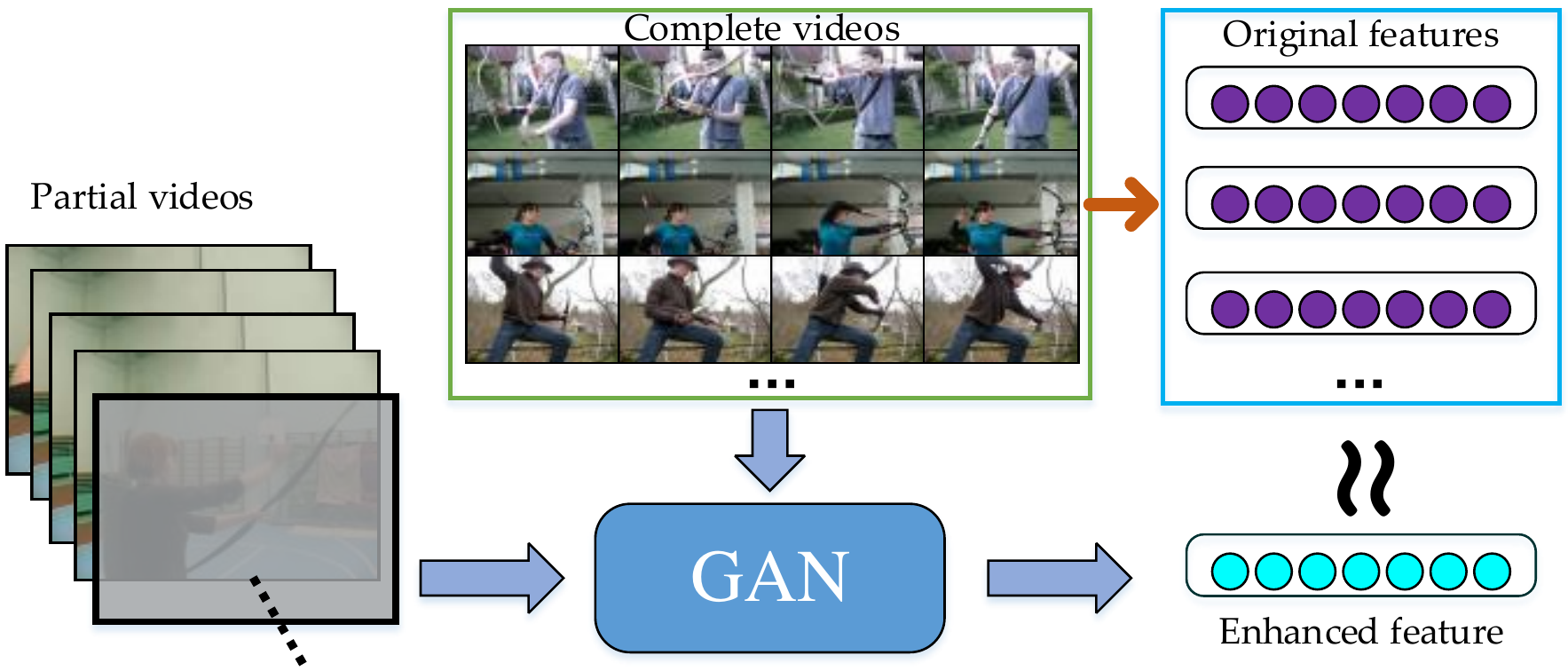}
  \caption{As shown in action classification, the feature representations of complete videos are discriminative while those of partial videos are of less information, which hurts the classification accuracy. In this work, we introduce the GAN model to enhance the feature representations from partial videos to be similar to the ones from complete videos, and thus improve the performance of action prediction task.}
\label{Fig-one}
\end{figure}

In this paper, we introduce the Generative Adversarial Network (GAN) for action prediction firstly. As illustrated in Fig. \ref{Fig-one}, the GAN aims to enhance the features of the partially observed videos to be similar to those of the complete videos, through learning to estimate residual error between features of the partially observed videos and complete ones. Similar to vanilla GAN, there are a generator network and a discriminator network that competes with each other. Concretely, the generator consists of a CNN and an LSTM, where the CNN is utilized for feature extraction and the LSTM is built upon it for residual error estimation. Then, the outputs of these subnetworks are added together to enrich the features of the incomplete videos and improves their discriminative power, obtaining the ``fake examples'' for the discriminator network; and the original features from complete videos are regarded as ``real examples''. The discriminator network is achieved by a fully-connected network, which is identical to vanilla GAN whose discriminator is trained to distinguish fake and real examples. Moreover, a perceptual network that equipped with traditional action recognition objective is employed, which ensures that the enhanced features retain discriminative characteristics between different action category and boosts the final action prediction accuracy.

To optimize the whole network, we follow the established training procedure for the generative adversarial network, \emph{i.e.} the generator and discriminator network are updated in an alternative manner to solve the min-max problem. In particular, the discriminator aims to assess whether the generated features from generator are indistinguishable from original features of complete videos; on the contrary, the generator tries to fool the discriminator with the enhanced features from incomplete videos. During training, the generator and discriminator act as competitors -- generator aims to generate indistinguishable features, whereas discriminators aims at distinguishing them. In addition, the action classification loss is passing to the generator network through the proposed perceptual network. After this adversarial training between those three networks, the enhanced features obtained from generator are similar to original features from complete videos and capable of providing high action prediction accuracy.

We conduct the experiments on UCF101, BIT and UT-Interaction datasets, which are widely used in action recognition and action prediction task. Our method consistently performs well on all datasets and outperforms the state-of-the-art methods, which demonstrates its superiority on action prediction. In summary, we make the following contributions:
\begin{itemize}
\item We introduce an end-to-end framework for action prediction task by extending the classical Generative Adversarial Network. It enhances the features from partially observed videos for achieving high action prediction accuracy.
\item A new generator tailored for sequential action prediction is proposed, which learns to estimate residual error between features of the partially observed videos and complete ones. Moreover, a perceptual network is employed to improve the discriminability of enhanced features of incomplete videos in conjunction with a discriminator that aims at differentiating fake and real examples.
\item Our method achieves significant improvements on publicly available datasets compared the state-of-the-art methods, especially for videos that less than 50\% portion of frames is observed.
\end{itemize}

\begin{figure*}
\begin{center}
   \includegraphics[width=0.9\linewidth]{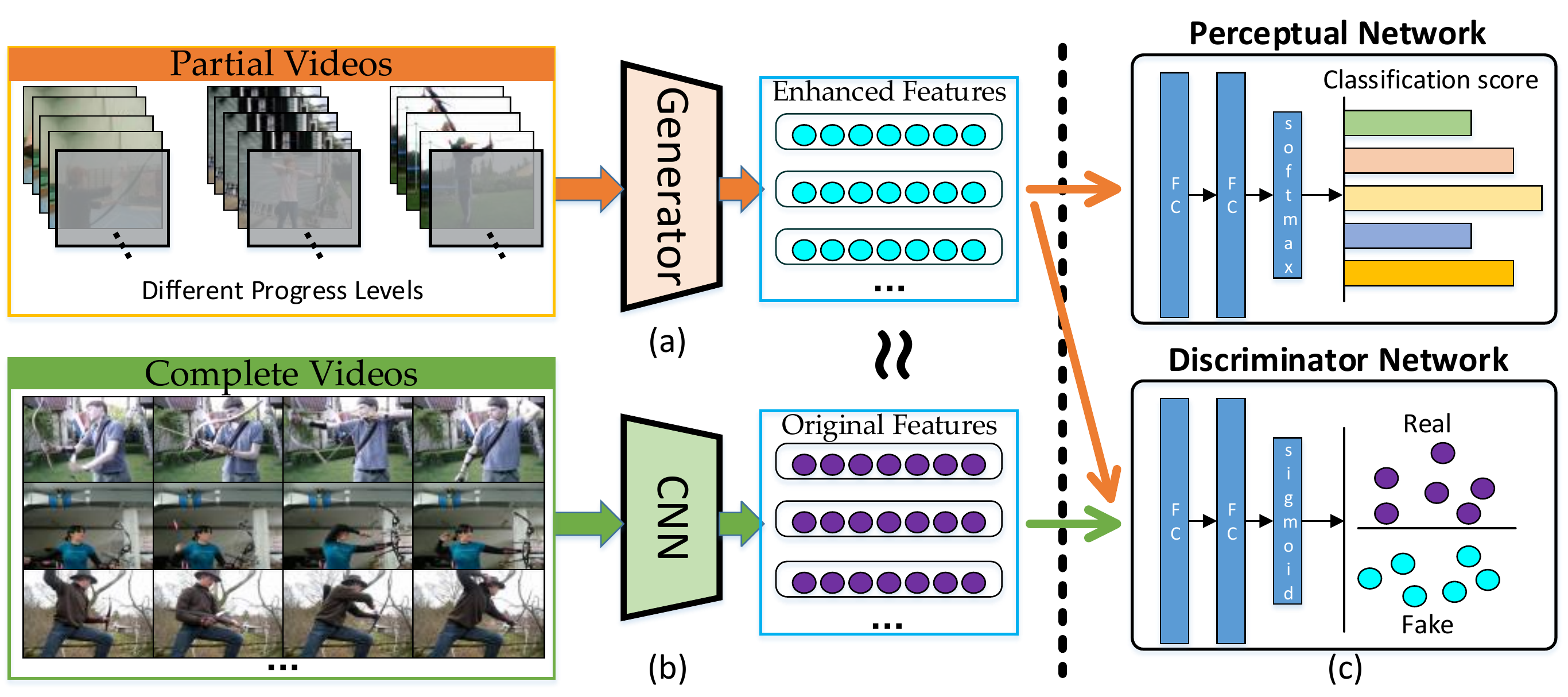}
\end{center}
     \caption{The overview of the proposed Generative Adversarial network. (a) The generator takes the partial videos as input and generates the enhanced features, which are regarded as fake examples for the discriminator.(b) A CNN is utilized to extracted original features from complete videos, which are regards as real examples. (c) The perceptual network consists of three fully connected layers followed by softmax activation, which is used to for action prediction based on the enhanced features. Similarly, the discriminator network consists of three fully connected layers followed by sigmoid activation, which is used to estimate the probability that the current input feature belongs to that of complete video. The whole network is end-to-end trainable using adversarial optimization framework.}
\label{Fig-two}
\end{figure*}

\section{Related Work}
Action prediction problem is an online version task of action recognition, which aims to infer the action label using fully observed videos. Therefore, there are many common techniques between these two tasks, and we review the works for action recognition at first. Previous works on action recognition can be roughly categorized into hand-crafted feature-based approaches \cite{laptev2005space,wang2011action,wu2014human,wang2013dense,ni2015motion,raptis2013poselet} and deep learning feature-based approaches \cite{simonyan2014two,yue2015beyond,karpathy2014large,wang2015action,feichtenhofer2016convolutional,du2015hierarchical}. Most popular hand-crafted features are dense trajectory and its improved version (improved trajectories) \cite{wang2011action}, which calculates dense point trajectories using optical flow and have demonstrated its effectiveness in action recognition. On the other hand, following the success of Convolutional Neural Networks (CNNs) on other computer vision tasks such as image classification \cite{krizhevsky2012imagenet}, object detection \cite{ren2015faster}, and sematic segmentation\cite{long2015fully}, deep learning method has been applied on action recognition successfully and achieved remarkable performance on public benchmark datasets. As a pioneer, Karpathy \emph{et al.} \cite{karpathy2014large} directly employ the pre-trained CNNs in extracting high-level feature on multiple frames of one video sequence. For exploiting motion information of video sequence, Karen Simonyan \emph{et al.} \cite{simonyan2014two} propose a novel two-stream CNN architecture that processes motion information with a separate CNN that is fed with optical flow, and gain a significant improvement on action recognition accuracy. Due to the promising performance of this two-stream architecture, many works \cite{wang2016temporal,sun2017lattice,zhang2016real} exploit different inputs and training strategy for two-stream architecture and achieve the state-of-the-art results.

Due to temporally incomplete videos, action prediction is more difficult and cannot employ action recognition in a straight way. For recognizing ongoing activities from streaming video, Ryoo \cite{ryoo2011human} represents an activity as an integral histogram of spatio-temporal features, efficiently modeling how feature distributions change over time. The action prediction methodology named dynamic bag-of-words is developed, which considers sequential nature of human activities while maintaining advantages of the bag-of-words to handle noisy observations. Then, Cao \emph{et al.} \cite{cao2013recognize} formulate the action prediction problem as a posterior-maximization formulation whose likelihood is calculated on each activity temporal stage using a sparse coding technique, which can better represents the activities with large intra-class variations. Moreover, An early event detector \cite{Hoai-DelaTorre-CVPR12} is proposed to localize the starting and ending frames of an incomplete event. For predicting future actions from a single frame in the challenging real-world scenarios, Lan \emph{et al.} \cite{lan2014hierarchical} introduce a novel representation to capture multiple levels of granularities in human movements. However, these works do not explicitly exploit an important prior of the video sequences, \emph{i.e.} discriminative action information is increasing when new observations are available and full observations contain all the useful information for action classification. Kong \emph{et al.} \cite{kong2014discriminative} harness this prior by embedding a label consistency constraint of temporal segments into the structured SVM objective, which considers multiple temporal scale dynamics of human actions and improves the performance of action prediction effectively. Ma \emph{et al.} \cite{ma2016learning} argue that the classification score of the correct activity category should be monotonically non-decreasing as the model observes more of the activity. They design a novel ranking losses to penalize the model on violation of such monotonicities, which are used together with classification loss in training of LSTM models. Recently, abundant sequential context information from complete action videos is explicitly exploited to enrich the feature representations of partial videos in \cite{kong2017deep}, which utilizes a deep architecture to improve the representation power of features and achieves impressive performance on several large scale action prediction datasets.

From a technical standpoint, our approach is based on the conditional GAN technique. GANs \cite{goodfellow2014generative} and conditional GANs \cite{mirza2014conditional} are popular formulations for learning generative models. For computer vision, GANs are originally introduced to generate images \cite{mathieu2015deep,denton2015deep}. In addition, conditional GANs are employed to learning a mapping from one manifold to another one, such as image style transfer \cite{li2016combining}, image inpainting \cite{yeh2016semantic} and image captioning \cite{chen2017show}. Moreover, Walker \emph{et al.} \cite{walker2017pose} exploit GAN to predict the future frames of the video in pixel space, which does not take the semantic action label into consideration. In recent work \cite{Zeng_2017_ICCV}, the generative adversarial imitation learning of Ho and Ermon \cite{ho2016generative} is utilized to forecast the intermediate representation of the future frames, and a linear-SVM is implemented to classify the action based on the anticipated representation. Despite its success in action prediction, it is a two-stage framework and not end-to-end trainable. Compared with this work, we extend the classical GAN framework and make it end-to-end trainable for action prediction task.

\section{Framework}
The goal of action prediction is to predict the action label $y$ of an action video $\rm{x}$ before the ongoing action execution ends. We follow the problem setup described in \cite{kong2017deep}. Given a fully observed video $\rm{x}$ of length $T$ (i.e. $T$ frames in total), we uniformly divide it into $K$ equal-length segments ($K = 10$ in this work), mimicking sequential video streaming in practical video surveillance system. Therefore, each segment of video $\rm{x}$ contains $\frac{T}{K}$ frames, and the $k$-th segment ($k \in \{ 1, \cdots ,K\}$) of the video ranges from the $\left[ {(k - 1) \times \frac{T}{K} + 1} \right]$-th frame to the $(\frac{{kT}}{K})$-th frame. Note that for different videos, the length $T$ might be different, and therefore the segments from different videos may have different lengths. Moreover, a temporally partial video $\rm{x}^{(k)}$ represents a temporal subsequence that contains the beginning $k$ out of $K$ segments of the video, and its progress level $g$ is $k: g = k$ and its observation ratio $r$ is $\frac{k}{K}: r = \frac{k}{K}$, where progress level $g$ of a partially observed video is defined as the number of observed segments that the video has and the observation ratio $r$ is the ratio of the number of frames in a partially observed video $\rm{x}^{(k)}$ to the number of frames in the full video $\rm{x}$.

Let $\mathcal{D} = \{ {{\rm{x}}_i},{y_i}\} _{i = 1}^N$ be the training data, where ${\rm{x}}_i$ is the $i$-th fully observed action video and ${y_i}$ is the corresponding action label. Moreover, partial observation with various progress levels are utilized to simulate sequential video streaming in this work, which is denoted as $\{ {\rm{x}}_i^{(k)}\} |_{k = 1}^K$. Note that ${\rm{x}}_i^{(K)}$ and ${\rm{x}}_i$ are the same full video: ${\rm{x}}_i^{(K)} = {\rm{x}}_i$. Firstly, a feature mapping function $G:{{\rm{x}}^{(k)}} \to {\bf{z}}$ is learned to extract feature of a partial video, which is achieved by the generator network in our GAN framework. Then, a prediction function $F:{\bf{z}} \to y$ that infers action label $y$ using the learned feature vector $\bf{z}$ is realized by the perceptual discriminator network.

\subsection{Overview}
As shown in \cite{kong2014discriminative,kong2017deep}, the discriminative power of partial observed video can be improved significantly by conveying the information from full videos. Inspired by this observation, we enhance the discriminative abilities of the features from a partial video by forcing them indistinguishable from the features from the full video. In this way, we believe that features from a partial video and corresponding full video are localized in the same manifold and capture the similar information about action category. Suppose ${\bf{z}}_i$ and ${\bf{g}}_i^{(k)}$ denotes the features from the full video ${\rm{x}}_i$ and the partial one ${\rm{x}}_i^{(k)}$, the discrepancy between ${\bf{z}}_i$ and ${\bf{g}}_i^{(k)}$ is minimized by an adversarial training mechanism between an sequential generator and an discriminator, where the discriminator learn to distinguish between ${\rm{x}}_i$ and $\{ {\rm{x}}_i^{(k)}\} |_{k = 1}^K$ while the generator aims to fool the discriminator by generated features from partial videos. Specifically, the learning objective for the proposed perceptual GAN model corresponds to a minimax two-player game, which is formulated as
\begin{equation}
\label{eq-gan}
\begin{split}
\mathcal{L}_{GAN}(G,D) &= {E_{{\bf{z}} \sim p({\bf{z}})}}\log D(\bf{z}) \\
 &+ {E_{{{\bf{g}}^{(k)}} \sim p({{\bf{g}}^{(k)}})}}[\log(1 - D({\bf{g}}^{(k)}))],
\end{split}
\end{equation}
where $G$ represents the sequential residual error generator that learns to generate the enhanced features conditioned on partial videos, and $D$ represents the perceptual discriminator that evaluates the \emph{quality} (the probability of a feature coming from the full video feature distribution $p({\bf{z}})$) of the input features. Moreover, ${\bf{g}}^{(k)}$ represents the enhanced features of partial videos that generated by the generator. The training procedure of $G$ is minimize this objective against an adversarial $D$ that tries to maximize it, i.e. ${G^*} = \arg {\min _G}{\max _D}{\mathcal{L}_{GAN}}(G,D)$.

To improve the discriminative ability of the generated feature for final action prediction, a more perceptual objective is mixed with GAN objective for optimize the generator. As shown in Fig. \ref{Fig-two}, a new subnetwork is built on the generator besides the discriminator, named perceptual network, which guides the generator to benefit the action prediction accuracy. Specifically, the standard categorial cross-entropy loss is utilized to provide feedback aboout action prediction precision to the generator, \emph{i.e.}
\begin{equation}
\label{eq-per}
{L_p}(G) = {E_{{{\bf{g}}^{(k)}},y \sim p({{\bf{g}}^{(k)}},y)}}[{L_{cls}}(y,P({{\bf{g}}^{(k)}}))],
\end{equation}
where $P$ represents the perceptual network that maps the generated features to final classification scores, $p({{\bf{g}}^{(k)}},y)$ represents the partial feature distribution with progress level $k$ and action label $y$. And $L_{cls}$ denotes the categorial cross-entropy loss function. Then, our final objective for the generator is
\begin{equation}
\label{eq-final-loss}
{G^*} = \arg \mathop {\min }\limits_G \mathop {\max }\limits_D {L_{GAN}}(G,D) + \lambda {L_p}(G).
\end{equation}

After adversarial training between the generator and the discriminator network, the perceptual network is also optimized for action classification based on the generated features. At inference, the generator and perceptual are concatenated to output a classification score vector for a partial video, and the discriminator network is not necessary anymore.

\begin{figure}[t]
  \centering
  \includegraphics[width=.45\textwidth]{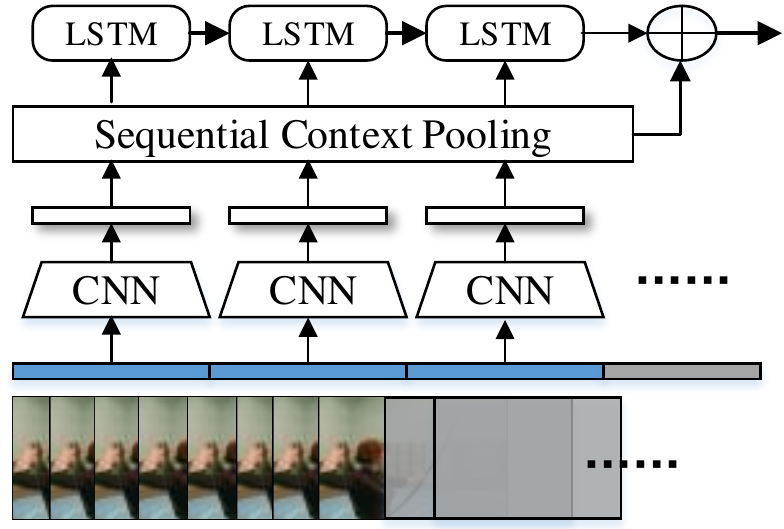}
  \caption{Details of the proposed sequential generator network.}\label{Fig-three}
\end{figure}

\subsection{Sequential Generator Network}
The generator network aims to generate enhanced features for partial videos to improve action prediction accuracy. To achieve this purpose, the generator is designed as a deep residual learning network that augments the features of partial videos to enhanced ones by adding residual error between features of the partially observed videos and complete ones, which is estimated with a tailored recurrent network.

As shown in Fig. \ref{Fig-three}, the image features are extracted from every segment of the partial videos using a typical convolution neural network (e.g. GoogLeNet), and then a two-layer LSTM network is adopted to learns the residual error between features of the partially observed videos and complete ones, which is the core of the generator. Instead of taking the individual image feature as the input for LSTM network, a sequential context pooling is operated on all observed frame features. Specifically, suppose ${\bf{f}}_{i}(i \in \{ 1, \cdots , k\})$ represents the individual segment feature of partial video $\rm{x}^{(k)}$, and then the sequential context pooling is formulated as follows:
\begin{equation}
\label{eq-sq-pool}
{{\bf{m}}_{i}} = \frac{1}{i}\sum\limits_{j = 1}^i {{\bf{f}}_{i}}.
\end{equation}
After this, the pooled features $\bf{m}_i$ capture more information of partial videos with different observation ratio, and the experimental results demonstrate its effectiveness. Then, these pooled features are fed into the LSTM network and the estimated residual error ${\bf{r}}^{(k)}$ is obtained at the last timestep. Finally, we combine the last step pooled feature and the estimated residual error via addition, \emph{i.e.} ${{\bf{g}}^{(k)}} = {{\bf{m}}_k} + {{\bf{r}}^{(k)}}$, where ${{\bf{g}}^{(k)}}$ represents the output feature of the generator network.
\subsection{Discriminator and Perceptual Network}
As shown in Fig. \ref{Fig-two}, the discriminator network consists of two fully-connected layers followed by a output layer with the sigmoid activation. Taking the enhanced features $\bf{g}^{(k)}$ from partial video and the original features $\bf{z}$ from full observed videos as input, the discriminator network is trained to differentiate them with following objective:
\begin{equation}
\label{eq-d-obj}
{D^*} = \arg \mathop {\max }\limits_D {L_{GAN}}(G,D).
\end{equation}
After training with this objective, the discriminator assesses the probabilities of the input coming from the full video feature distribution $p(\bf{z})$. Similarly, the perceptual network is composed of two fully-connected layers followed by a classification layer with softmax activation, which maps the generated feature to the action classification scores. Suppose $s \in {R^C}$ represents the output vector from the last classification layer, where $C$ is the number of action category, and $y \in {R^C}$ is the one-hot encoded vector of groundtruth action label. Then, the objective loss function of the perceptual network is formulated as
\begin{equation}
\label{eq-p-obj}
{L_{cls}}(y,P({{\bf{g}}^{(k)}})) =  - \sum\limits_{i = 1}^C {{y_i}\log {s_i}}.
\end{equation}
Using this loss function, the discriminative information between different action category is back-propagated to the generator network through the perceptual network, which forces the generated feature must be beneficial for action prediction. The output units number of the first two fully-connected layers for these two network are 4096 and 1024 respectively.

\subsection{Optimization and Inference}
To optimize our network, a two-stage training procedure is designed for the whole network. First, we concatenate the feature extraction CNN and the perceptual network for action classification using full videos, and the feature extraction CNN is fine-tuned to generate the discriminative feature representation for action classification by minimizing typical classification objective. After this supervised pre-training, we believe that the feature representations from full videos capture the most important information of the whole video for action classification and are regarded as real examples in adversarial training stage. Then, the whole network is trained to predict the action labels for partial videos by adversarial training framework. Specifically, we follow the standard approach from \cite{goodfellow2014generative}, \emph{i.e.} we alternative between one gradient descent step on the generator using Eq. \ref{eq-final-loss}, then three steps on the discriminator using Eq. \ref{eq-d-obj}.

At inference time, we just run the generator and perceptual network with partial videos in the same manner as the training phase, and the discriminator network is abandoned during the test phase. Considering computation issue, we just sample 1 frame from each segment of the video, whose effectiveness is demonstrated from experimental results. Moreover, inspired by two-stream architecture in action classification field, a sibling network is trained with optical images and utilized to improve final performance.

\section{Experiments}
\subsection{Dataset and Experiment Setup}
We evaluate our approach on three datasets: UCF101 dataset \cite{soomro2012ucf101}, BIT-Interaction dataset \cite{KongECCV2012} and UT-Interaction dataset \cite{ryoo2009spatio}, which are widely used in action classification field. UCF101 dataset consists of 13320 action videos in 101 categories and BIT dataset consists of 8 classes of human interactions, with 50 videos per class. The UT-dataset contains a total of 120 videos of 6 classes of human interactions, which is collected from real surveillance videos. Although each video in UCF101 datasets only contains a single action category in these datasets, the action instance may appear anywhere with different duration in the video, which makes the action prediction more challenging. Moreover, It should be noted that $N$ videos will be $10N$ videos to action prediction approaches due to the modeling of 10 progress levels.

UCF101 dataset has three split settings to separate the dataset into training and testing videos. As for BIT dataset, we follow the experiment settings in \cite{kong2014discriminative}, and use the first 34 videos in each class for training and use the remaining for testing. In order for a fair comparison with other reported numbers on UT-dataset, we follow the experiment settings as in \cite{ryoo2011human}.

\begin{figure*}
\centering
\subfigure[UCF101 dataset]{
\begin{minipage}[t]{0.3\textwidth}
\centering
\includegraphics[width=5cm]{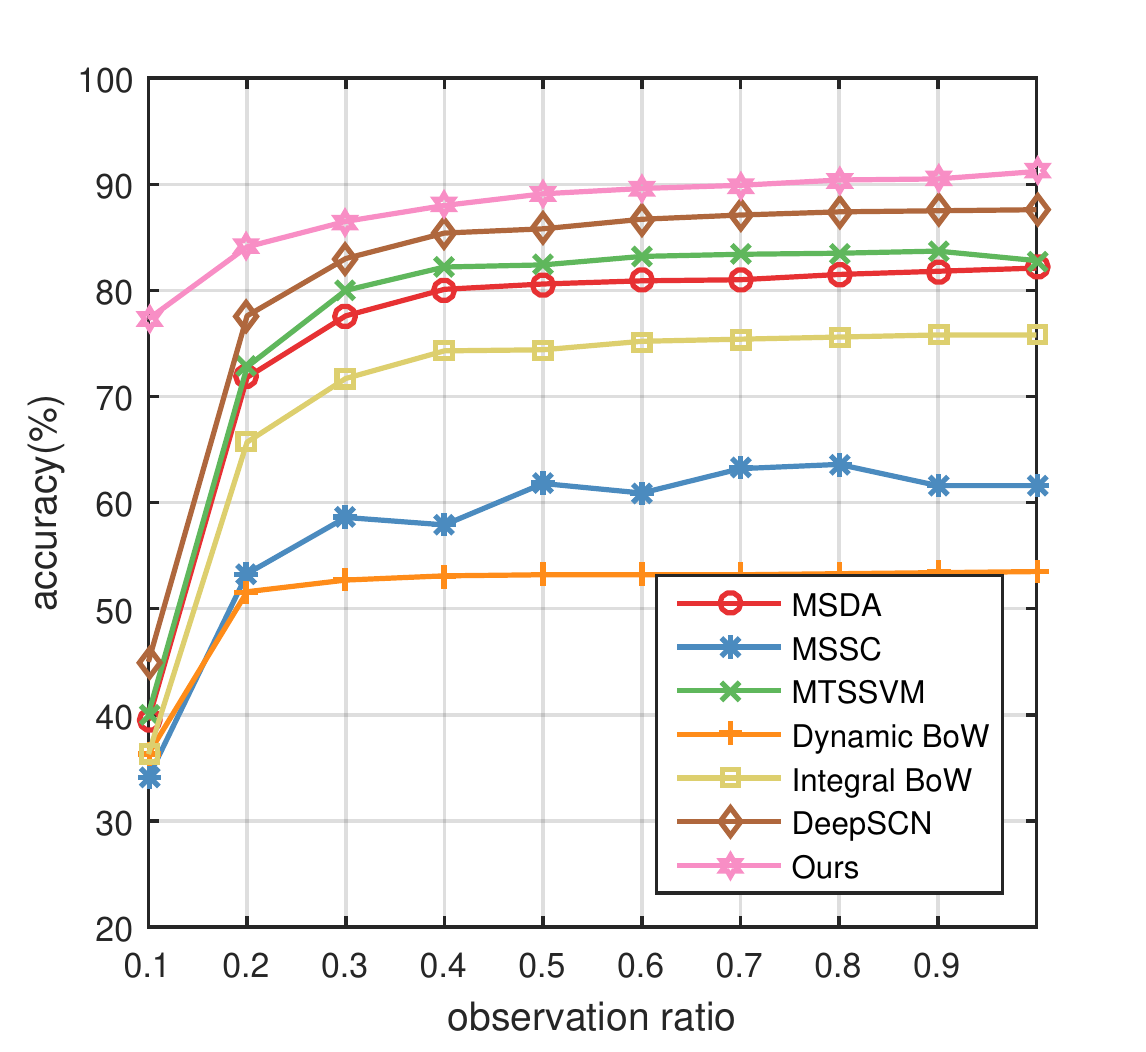}
\end{minipage}
}
\subfigure[BIT dataset]{
\begin{minipage}[t]{0.3\textwidth}
\centering
\includegraphics[width=5cm]{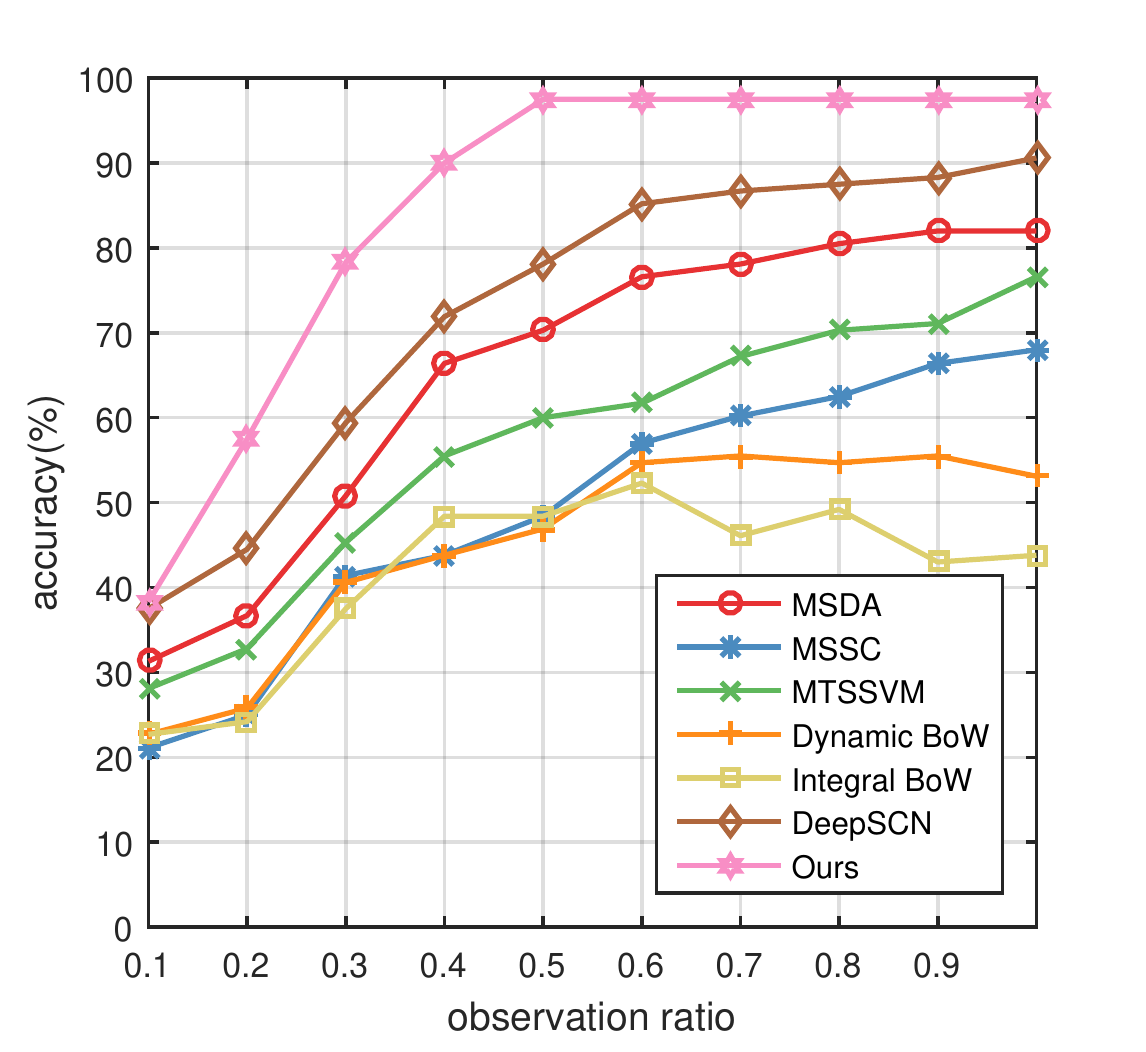}
\end{minipage}
}
\subfigure[UT dataset]{
\begin{minipage}[t]{0.3\textwidth}
\centering
\includegraphics[width=5cm]{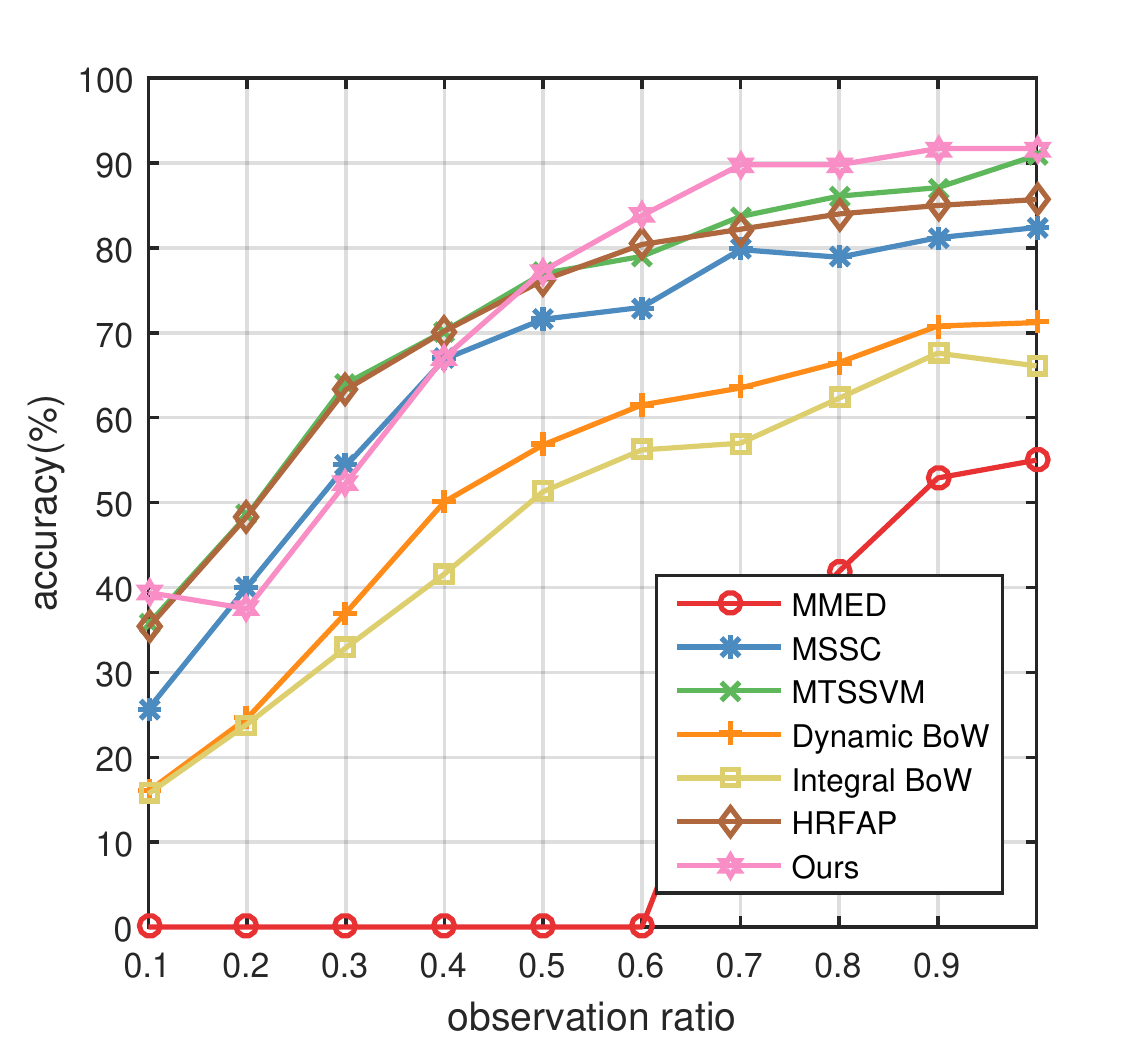}
\end{minipage}
}
\caption{Prediction results on UCF101, BIT and UT dataset. Note that these prediction approaches are tested with different observation ratios, \emph{i.e.} from 0.1 to 1.0.}
\label{Fig-ALL}
\end{figure*}

\subsection{Implementation Details}
\subsubsection{Two-Stream Network Architecture}
Inspired by the grate success of two-stream neural network in action recognition, we also explore optical flow modality to enhance the discriminative power of the overall framework. Specifically, there are two networks with identical architecture, which are trained independently. The only difference between these two networks is the input data, \emph{i.e.} RGB images for one network and optical flow images for the another one, denoted as Net$_{rgb}$ and Net$_{of}$ respectively. Moreover, there is no fusion component between these two networks, which are trained with Eq. \ref{eq-final-loss} separately. When testing, the final classification scores are obtained by summing the scores from these two networks.

We choose the Inception with Batch Normalization (BN-Inception) as the building block of our CNN feature extractor for RGB and optical flow images. The CNN feature extractor is pre-trained on ImageNet \cite{krizhevsky2012imagenet} and finetuned with the corresponding action prediction task. The sequential generator, discriminator, and perceptual network are all trained from scratch. The parameter $K$ is set to 10, and we sample only one frame in each segment to represent this segment, whose effectiveness has been demonstrated in action classification task \cite{wang2016temporal}. In other words, the segment feature ${\bf{f}}_i$ is obtained by passing one frame into CNN feature extractor and its dimension is 1024. Due to 10 segments in each video, our two-layer LSTM network in the sequential generator contains 10 unrolled time steps and is trained using back-propagation through time (BPTT), and its hidden state dimension is set to 1024. In the first stage, we use mini-batch stochastic gradient descent (SGD) to training the network with an initial learning rate of 0.001, a momentum of 0.9, and a weight decay $5 \times {10^{ - 4}}$. We decay the learning rate to its $1/10$ every 4500 iterations, and the first training procedure stops at 18000 iterations. In the adversarial training stage, the networks are all optimized with ADAM solver with an initial learning rate of 0.0001 and the same momentum and weight decay. The parameter $\lambda$ in Eq. \ref{eq-final-loss} is set to 1 and the batch size is set to 64 in both stages.

\begin{table*}[htbp]
\centering
\caption{The actions ratio that achieve $\ge$60\%, $\ge$80\%, and $\ge$90\% accuracy when 10\%, 50\%, and 100\% frames are observed.}
\begin{tabular}{|c|c|c|c||c|c|c||c|c|c|}
\hline
\multirow{2}{*}{Auccary} &
\multicolumn{3}{c||} {UCF101} &
\multicolumn{3}{c||} {BIT} &
\multicolumn{3}{c|} {UT}\\
\cline{2-10}
&0.1 &0.5 &1.0 &0.1 &0.5 &1.0 &0.1 &0.5 &1.0 \\
\hline
\hline
$\ge$60\% &91.09 &98.02 &99.01 &12.52 &100.00 &100.00 &25.00 &75.00 &100.00\\
$\ge$80\% &71.29 &85.15 &87.13 &0.00 &100.00 &100.00 &19.45 &58.34 &83.34\\
$\ge$90\% &35.64 &64.36 &69.31 &0.00 &87.55 &100.00 &8.34 &25.00 &50.00\\
\hline
\end{tabular}\label{Table-action-ratio}
\end{table*}

\subsection{Performance Comparison}
For evaluating the overall performance, we compare the proposed method with Dynamic BoW (DBoW) and Integral BoW (IBoW) \cite{ryoo2011human}, MMED \cite{Hoai-DelaTorre-CVPR12}, MSSC \cite{cao2013recognize}, MTSSVM \cite{kong2014discriminative}, HRFAP \cite{lan2014hierarchical} and DeepSCN \cite{kong2017deep}. Moreover, SVM with marginalized stacked autoencoder (MSDA) \cite{chen2012marginalized} is used for comparison. The experimental results are summarized in Fig. \ref{Fig-ALL}. Overall, the proposed method outperforms the previous state-of-the-art method of Kong \emph{et al.} \cite{kong2017deep} in terms of average accuracy 87.7\% vs 81.3\% on UCF101 dataset and 84.9\% vs 73.0\% on BIT dataset. As for UT dataset, our method achieve 72.1\% accuracy, which is comparable with the current state-of-the-art performance 72.2\% \cite{kong2014discriminative}.

Specifically, our approach makes a large improvement, \emph{i.e.}, 27.8\% in average accuracy when only the beginning 10\% portion of the video is observed on UCF101, demonstrating its superiority in predicting action instantly. Moreover, the performance of our method at observation ratio 0.3 is already higher than the best performance of all the other comparison methods, suggesting the advantage of transferring discriminative information from complete videos. Note that our method achieves an impressive 90.4\% when only 50\% frames are observed, which is comparable with the performance when the whole video is observed, showing the effectiveness of its early action prediction ability. In addition, as shown in Fig. \ref{Fig-ALL}, the performance of our method is increasing persistently along with the video processing, showing the benefits of the enhanced feature from the generator.

As for BIT dataset, there is no accuracy improvement compared with the previous state-of-the-art method of Kong \emph{et al.} \cite{kong2017deep} when observation ratio is 0.1. The reason behind this is that the BIT dataset consists of various human interaction actions (such as boxing, handshake, high-five, and so on), whose appearance are similar in the very beginning part, like two people standing together. After receiving more information from the video, the accuracy is improved rapidly and achieves 78.5\% at observation ratio $r=0.3$, 18.3\% higher than the runner-up DeepSCN method. At $r=0.5$, our method achieves a remarkable result of 97.5\%, which makes a significant improvement over the previous best performance even at observation ratio $r=1.0$. Moreover, there is no accuracy vibration when more frames are observed, which demonstrates our method is capable of extracting most discriminative information while ignoring noise.

Results on UT dataset show that our method obtains a comparable performance on the average. Our method achieves 77.3\% at observation ratio r=0.5, which is 0.3\% higher than the MTSSVM. After exploring the UT dataset, we find that almost all actions start when 50\% frames are observed, which also explains the inferior accuracy of our method when observation ratio is 0.2-0.4. Our method outperforms all the other comparison methods when observation ratio is greater than 0.5, demonstrating the superiority of our method.

\begin{figure*}
\begin{center}
   \includegraphics[width=0.9\linewidth]{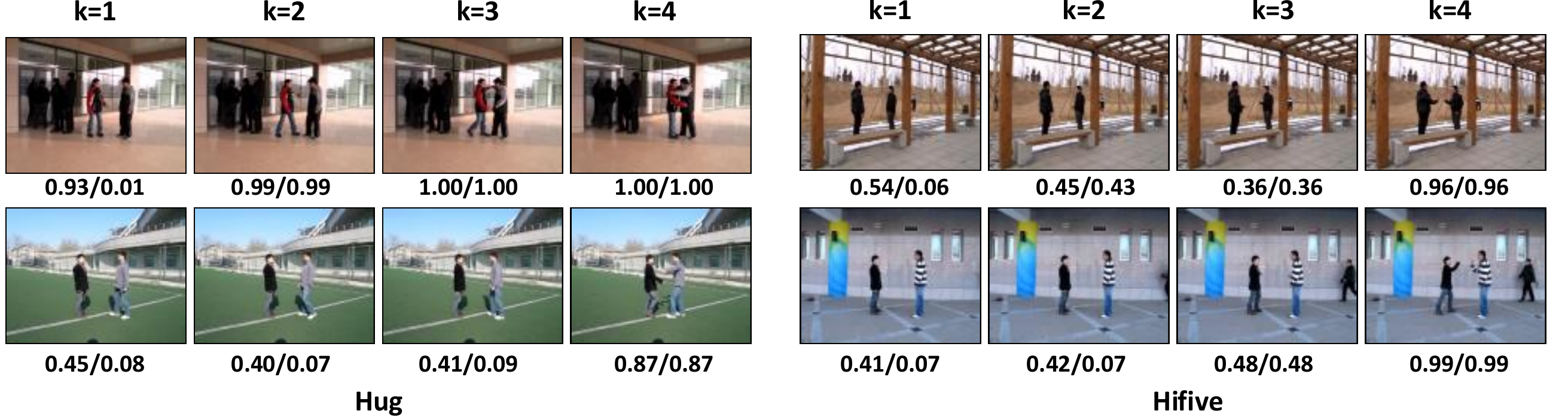}
\end{center}
     \caption{Early action prediction results of the proposed method on BIT dataset. Note that the proposed method can correctly prediction action label once the discriminative patterns between different action categories appear.}
\label{Fig-BIT-Vis}
\end{figure*}

\subsection{Early Prediction Analysis}
Action prediction aims at inferring action label as soon as possible, and this real-time requirement is crucial for practical application, such as video surveillance system. However, the discriminative patterns of actions may occur at different parts in different action video, that is, some action categories can be classified correctly after only observing the beginning few frames. In contrast, some actions are ambiguous in the beginning of the video, which are predictable only when the entire action execution is observed. We analyze the action prediction performance on different action categories at observation ratio $r=0.5$, and we denote the action prediction only using 50\% frames of a video as early prediction. Moreover, we also study the performance at observation ratio $r=0.1$ and $r=1.0$, which are called instant prediction and late prediction respectively.

Table. \ref{Table-action-ratio} shows that how many action categories achieve accuracy 60\%, 80\%, and 90\% at different observation ratio. To be specific, there are 71.29\% actions in 101 actions achieving $\ge 80\%$ accuracy when only the beginning 10\% frames are observed, which means $\ge 80\%$ test videos of these action categories have been classified correctly, demonstrating the superiority of our method in early prediction. Moreover, 35.64\% actions achieves impressive $\ge 90\%$ prediction results at observation ratio $r=0.1$. For more detailed elaboration, we list the 5 action categories that achieve the top 5 accuracy at observation ratio $r=0.1$, $r=0.5$ and $r=1.0$ in Table. \ref{Table-action-ratio}. We can see that actions ``PlayingDhol'' and ``BasketballDunk'' achieve impressive prediction accuracy, \emph{i.e.}, 99.19\% and 99.17\% respectively, which means almost all test videos from these two categories are classified correctly after observing one frame from the first segment of the video. In fact, there are 36 actions out of 101 actions achieve $\ge 90\%$ prediction accuracy at observation ratio $r=0.1$, demonstrating the advantage of instant action prediction.

As shown in Table. \ref{Table-action-ratio}, the overall prediction accuracy is improved significantly with more frames being observed. 64.36\% actions achieve $\ge 90\%$ accuracy at observation ratio $r=0.5$, which is comparable to the results when all frames of video are available. This suggests that the discriminative information of most actions occurs in the first half part of the video, and our approach is efficient at discovering and representing these discriminative patterns. We also list the top 5 action categories at observation ratio $r=0.5$ in Table. \ref{Table-action-categories}. Results show that all these 5 actions achieve 100\% prediction accuracy after observing 50\% frames, which is impressive because the actions are totally uncomplete at observation ratio $r=0.5$. Furthermore, action ``VolleyballSpinking''  and ``PommelHorse'' are not top 5 actions at observation ratio $r=0.1$ and $2-3\%$ accuracy is improved on other actions, which indicates that the discrimination of actions is strengthened when more frames are observed. Overall, our method can discover the discriminative patterns of most action categories after observing the beginning 50\% frames of the videos and predict the correct action label for it. On BIT dataset, as shown in Table. \ref{Table-action-ratio}, the advantage of our method for early prediction is more prominent, and we analyze it by a visualization way in next section. As for UT dataset, similar results are obtained, demonstrating the robustness of our method.
\begin{table}[htbp]
\centering
\caption{Top 5 instant and early prediction action categories on UCF101 dataset.}
\small
\begin{tabular}{|c||c|}
\hline
Instant Prediction &Early Prediction\\
\hline
\hline
PlayingDhol (99.19\%) &\textbf{VolleyballSpinking} (100\%)\\
BasketballDunk (99.17\%) &\textbf{PommelHorse} (100\%)\\
Billiards (98.41\%) &Billiards (100\%)\\
ParallelBars (98.09\%) &PlayingDhol (100\%)\\
HorseRace (97.98\%) &HorseRace (100\%)\\
\hline
\end{tabular}\label{Table-action-categories}
\end{table}

\subsection{Early Prediction Visualization}
Several test examples from BIT dataset are shown in Fig. \ref{Fig-BIT-Vis}. Due to the page limit, we select two test videos from action ``Hug'' and ``Hifive'' respectively, and show the frames from beginning four segments of each video. The corresponding action prediction scores are shown at the bottom of frames in ``\emph{max score/true class score}'' formation. In this way, we can clearly see how the proposed method making more accurate prediction when more frames being observed.

Specifically, for first test video from action ``Hug'', the action prediction score for action ``Hug'' is 0.01 while 0.93 for action ``Pat'' when only one frame is observed; and the action prediction score from second test video show similar results when progress level $k=1$, \emph{i.e.}, 0.45 action ``Handshake'' for and 0.08 for action ``Hug''. The reason behind this is that the videos from different actions always start with the similar scene that two people is standing close and looking each other, such as ``Handshake'', ``Hug'', and ``Hifive''. The video examples for action ``Hifive'' reveal this problem in a more serious way. The class scores for action ``Hifive'' are not the maximum when progress level $k=1$ and $k=2$. Although the action prediction is correct when progress level $k=3$, its class score is 0.36 and 0.48, which indicates this prediction results are not reliable. These results are reasonable because there is no discriminative information about action ``Hifive'' provided from these frames.

On the other hand, our method makes the correct predictions once the discriminative patterns appear. As shown in Fig. \ref{Fig-BIT-Vis}, the frames from the second and third segment of first video capture the beginning posture of action ``Hug'', and the high prediction scores are obtained by our approach, \emph{i.e.}, 0.99 and 1.00. What is more, our method is not restricted to a specific progress level and the correct prediction is made whenever the discriminative patterns appear. For example, the confident action prediction results are produced at progress level $k=4$ for two test video from action ``Hifive'' and second test video from action ``Hug'', suggesting the superiority of our method in early action prediction.

\begin{table*}[htbp]
\centering
\caption{Comparisons of early prediction performance with several variants of the proposed method on BIT dataset.}
\begin{tabular}{c||c|c|c|c||c|c|c|c}
\hline
\multirow{2}*{Method} &
\multicolumn{4}{c||}{Net$_{rgb}$} &
\multicolumn{4}{c}{Net$_{of}$} \\
\cline{2-9}
&0.1 &0.3 &0.5 &Avg. &0.1 &0.3 &0.5 &Avg. \\
\hline
\hline
SCP &14.17\% &15.83\% &14.17\% &14.72\% &13.33\% &15.00\% &20.83\% &16.39\%\\
LSTM &15.83\% &21.67\% &46.67\% &28.06\% &29.17\% &64.17\% &79.17\% &57.50\%\\
LSTM + SCP &15.00\% &24.17\% &40.00\% &26.39\% &34.17\% &70.83\% &83.33\% &62.78\%\\
Ours &15.83\% &42.50\% &77.50\% &45.28\% &38.33\% &76.67\% &92.5\% &69.17\%\\
\hline
\end{tabular}\label{Table-Features}
\end{table*}

\begin{table*}[htbp]
\centering
\caption{Comparison of early prediction performance by different training configuration on BIT dataset.}
\begin{tabular}{c|ccc||c|ccc}
\hline
Net$_{rgb}$ &0.1 &0.3 &0.5 &Net$_{of}$ &0.1 &0.3 &0.5\\
\hline
\hline
SCP\_SP &10.83\% &11.67\% &15.83\% &SCP\_SP &11.67\% &13.33\% &16.67\%\\
SCP\_SP\_AD &14.17\% &15.83\% &14.17\% &SCP\_SP\_AD &13.33\% &15.00\% &20.83\%\\
\hline
LSTM\_SP &8.33\% &11.67\% &21.67\% &LSTM\_SP &10.83\% &13.33\% &22.50\%\\
LSTM\_SP\_AD &15.83\% &21.67\% &46.67\% &LSTM\_SP\_AD &29.17\% &64.17\% &79.17\%\\
\hline
SCP+LSTM\_SP &7.50\% &12.50\% &23.33\% &SCP+LSTM\_SP &25.83\% &50.00\% &69.17\%\\
SCP+LSTM\_SP\_AD &15.00\% &24.17\% &40.00\% &SCP+LSTM\_SP\_AD &34.17\% &70.83\% &83.33\%\\
\hline
Ours\_SP &12.50\% &33.33\% &68.33\% &Ours\_SP &26.67\% &64.17\% &88.33\%\\
Ours\_SP\_AD &15.83\% &42.50\% &77.50\% &Ours\_SP\_AD &38.33\% &76.67\% &92.50\%\\
\hline
\end{tabular}\label{Table-AD}
\end{table*}

\subsection{Ablation Studies}
\subsubsection{The Effectiveness of Enhanced Features}
For verifying the superiority of the enhanced feature in early action prediction, we compare our method with several variants baseline, including applying high-level image feature with sequential context pooling, modeling the sequential information by using LSTM, and combining LSTM with sequential context pooling. All these variants are implemented based on the same CNN and the perceptual network with end-to-end training and tested on the BIT dataset. Moreover, for demonstrating the superiority of our method on different input modalities, we also test these variants with RGB and optical flow image separately. As shown in Table. \ref{Table-Features}, ``SCP'' indicates the model trained by applying high-level image feature through sequential context pooling. Net$_{rgb}$ (trained with RGB images) outperforms this baseline by 30.56\% in average accuracy when observation ratio is 0.5, which validates that our method can generate more discriminative feature after observing less than 50\% frames of the complete videos. Similarly, Net$_{of}$ (trained with optical flow image) achieve 69.17\% when 50\% frames are observed, 52.78\% higher than the SCP baseline. ``LSTM'' represents the model trained by modeling the sequential information of the video by using typical LSTM. ``LSTM + SCP'' represents the model trained by feeding the sequential context pooling results into LSTM network. Note that both Net$_{rgb}$ and Net$_{of}$ outperforms these variants in early action prediction, demonstrating the superiority of our method in modeling sequential information.

\subsubsection{The Effectiveness of Adversarial Training}
The whole network proposed in this work is optimized through two-stage training procedure, \emph{i.e.}, supervised pre-training and adversarial training. For demonstrating the necessity of this training framework, we report the performance of our model with different training configuration in Table \ref{Table-AD}. In addition, we also apply our two-stage training procedure to several variants baseline, which are analysed in previous section. ``*\_SP'' indicates the model of training the generator and perceptual network with typical classification objective. ``*\_AD'' indicates the model of training the generator, discriminator, and perceptual network through adversarial training without supervised pre-training, which is not convergent in our experiments. Moreover, ``*\_SP\_AD'' represents the model trained using our two-stage procedure. By comparing ``*\_SP\_AD'' with ``*\_SP'' in different baseline setting, we can find that considerable improvements after adversarial training. To be specific, with RGB image input, SCP gains 1.95\% improvement in average accuracy. LSTM obtains the largest improvement with optical flow images. Moreover, the performance of the proposed method is boosted by 7.29\% and 9.45\% in averages respectively. These results shows that our method can improve its performance in early action prediction by enhancing the discriminative ability of the features from the generator through adversarial training.

\section{Conclusion}
In this paper, we introduce the generative adversarial network to address the challenging problem of early action prediction firstly. The proposed method enhances the features from partial videos by conveying discriminative information from corresponding complete videos, which is achieved by alternatively optimized generator network and discriminator network. The generator learns to enhance features from partial videos by estimating additive residual error between them and original features from complete videos. The discriminator aims at differentiating the enhanced features from the generator and original features from complete videos. Moreover, a perceptual network is employed to improve the discriminability of enhanced feature by back-propagating the action class information to the generator. Competition between these networks encourages the generator generates more discriminative and informative features for partial videos, thus improving action prediction performance. Extensive experiments have demonstrated the superiority of the proposed method in action prediction, especially early action prediction.

\ifCLASSOPTIONcaptionsoff
  \newpage
\fi

\bibliographystyle{IEEEtran}
\bibliography{access-paper}

\begin{IEEEbiography}[{\includegraphics[width=1in,height=1.25in,clip,keepaspectratio]{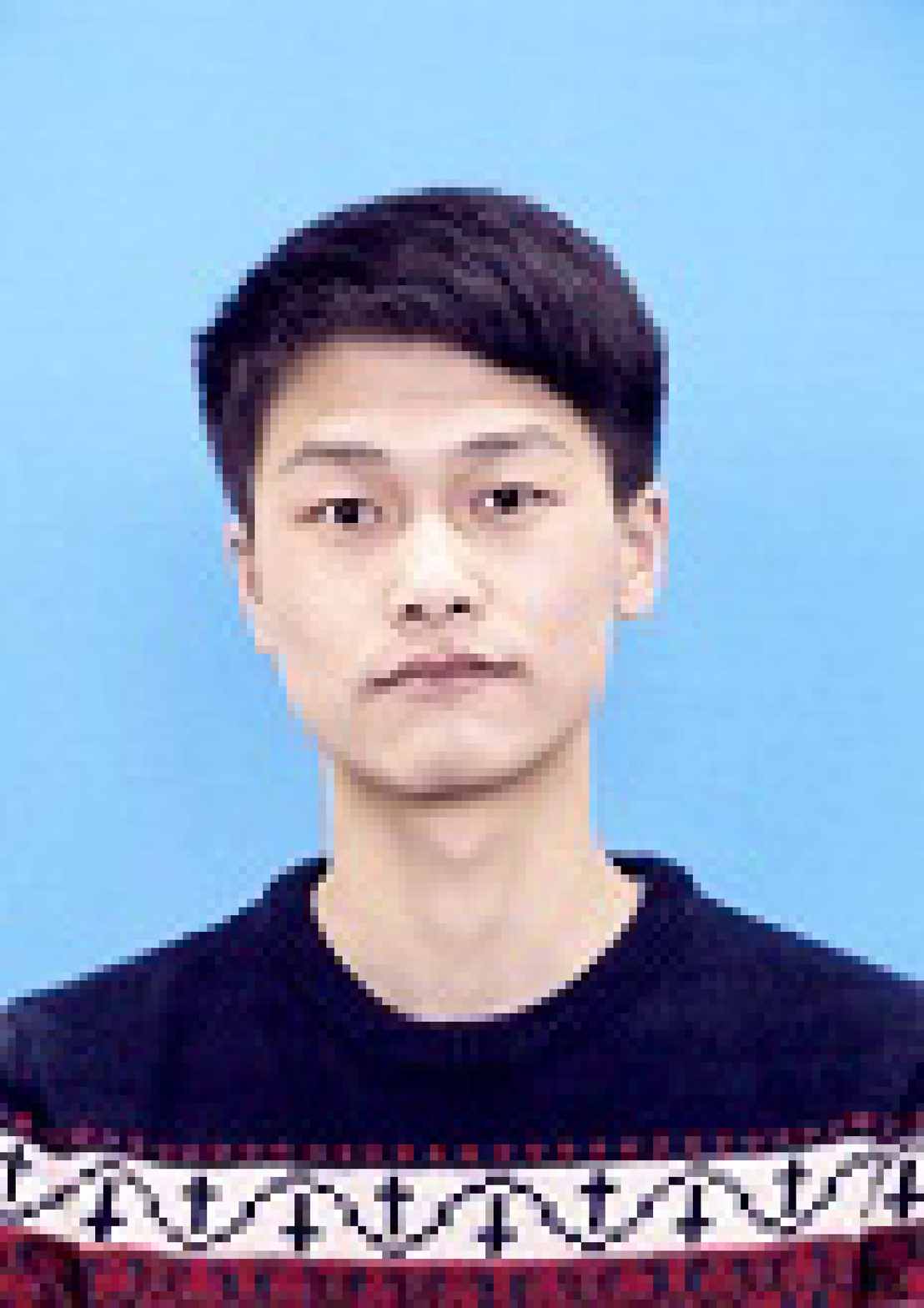}}]{Dong Wang} received the B.E. degree in Computer Sciense and Technology from the Northwestern Polytechnical University, Xi'an 710072, Shaanxi, P. R. China, in 2015. He is currently pursuing the Ph.D. degree from the Center for Optical Imagery Analysis and Learning, Northwestern Polytechnical University, Xi¡¯an, China. His research interests include computer vision and pattern recognition.
\end{IEEEbiography}

\begin{IEEEbiography}[{\includegraphics[width=1in,height=1.25in,clip,keepaspectratio]{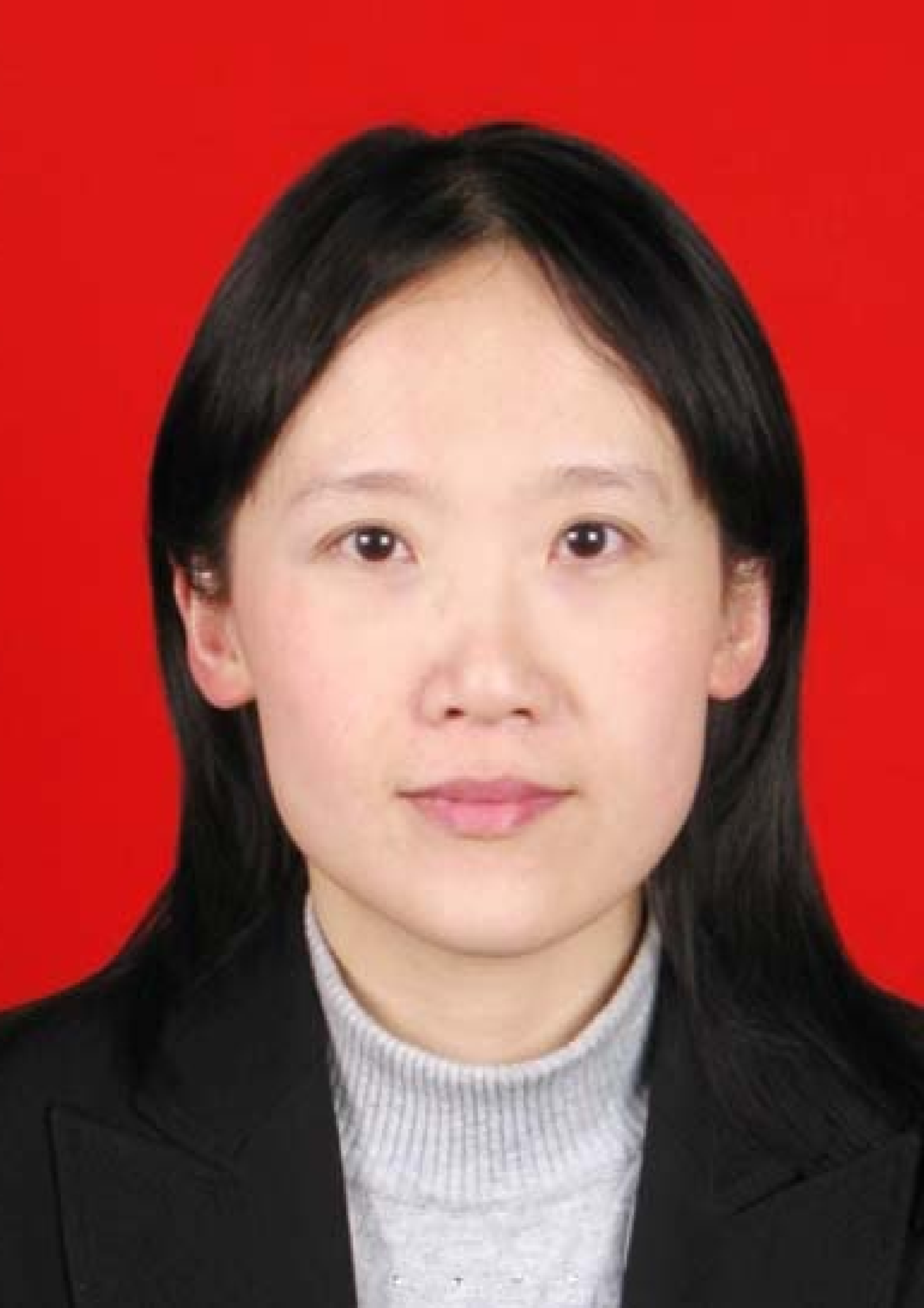}}]{Yuan Yuan} (M'05--SM'09)  is currently a Full
Professor with the School of Computer Science
and the Center for OPTical IMagery Analysis and
Learning, Northwestern Polytechnical University,
Xi'an, China. She has authored or co-authored
over 150 papers, including about 100 in reputable
journals, such as the IEEE TRANSACTIONS AND
PATTERN RECOGNITION, and also conference
papers in CVPR, BMVC, ICIP, and ICASSP. Her
current research interests include visual information
processing and image/video content analysis.
\end{IEEEbiography}

\begin{IEEEbiography}[{\includegraphics[width=1in,height=1.25in,clip,keepaspectratio]{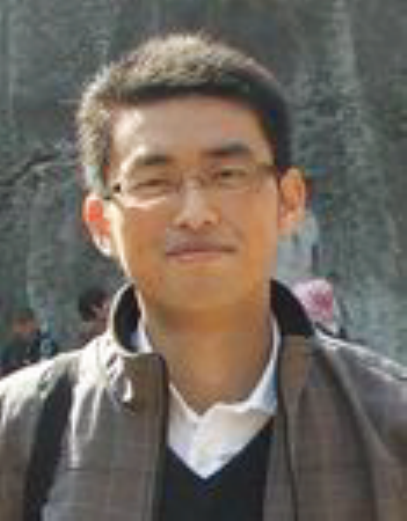}}]{Qi Wang} (M'15-SM'15) received the B.E. degree in automation and the Ph.D. degree in pattern recognition and intelligent systems from the University of Science and Technology of China, Hefei, China, in 2005  and 2010, respectively.  He is currently a Professor with the School of Computer Science and with the Center for OPTical IMagery Analysis and Learning (OPTIMAL), Northwestern Polytechnical University, Xi'an, China. His research interests include computer vision and pattern recognition.
\end{IEEEbiography}

\EOD

\end{document}